# Monocular Visual Analysis for Electronic Line Calling of Tennis Games

Yuanzhou Chen[1], Shaobo Cai[2(✉)], Yuxin Wang[2], and Junchi Yan[2]

[1] YK Pao School, Shanghai 200003, China
{s17680}@ykpaoschool.cn
[2] Shanghai Jiao Tong University, Shanghai 200240, China
{582875593,wyx-0279,yanjunchi}@sjtu.edu.cn

**Abstract.** Electronic Line Calling (ELC) is an auxiliary referee system used for tennis matches based on binocular vision technology. While ELC has been widely used, there are still many problems, such as complex installation and maintenance, high cost and etc. We propose a monocular vision technology based ELC method. The method has the following steps: (1) locate the tennis ball's trajectory. We propose a multistage tennis ball positioning approach combining background subtraction and color area filtering. (2) Then We propose a bouncing point prediction method by minimizing the fitting loss of the uncertain point; (3) Finally, we find out whether the bouncing point of the ball is out of bounds or not according to the relative position between the bouncing point and the court side line in the two-dimensional image.

We collected and marked 394 samples, we achieved 99.4% accuracy, and among the 11 samples with bouncing points near the line, we obtained 81.8% accuracy. The experimental results show that our method is feasible to judge if a ball is out of the court with monocular vision and significantly reduce complex installation and costs of ELC system with binocular vision.

## 1 Introduction

### 1.1 Research Background

Hawk-eye system is an instant replay system [1]. It can track and record the three-dimensional motion trajectory of the sphere through high-speed and high-quality camera and computer vision technology, predict the bouncing point according to the three-dimensional motion trajectory. It can also replay the virtual three-dimensional image in form of dynamic images and show the result to the audience and players. The hawk-eye system can display the game from multiple angles, overcome many restrictions of human judgment, and improve the accuracy of referee judgments and the fairness of the game. Furthermore, the hawk-eye system can help analyze the skills of the players, collect statistics, provide instruction for teaching ball sports. The hawk-eye system is not only used in tennis, but also in football, cricket, basketball, badminton, rugby, baseball, volleyball, billiards and other sports.

The first hawk-eye system was developed by Hawk-Eye Innovations. The system was initially implemented in 2001 and was used for TV broadcasts for cricket competitions. The hawk-eye system started to act as an auxiliary judgment in a variety of sporting competitions in a few years. In 2005, the Electronic Line Calling (ELC) technology in the Hawk-Eye system was recognized by the International Tennis League (ITF). In 2006, the US Tennis Open became the first grand slam that introduced a hawk-eye system. China Open also followed the use of ELC in the same year. In 2009, the hawk-eye system was officially introduced to cricket games. The goal-line technology used in Premier League was based on hawk-eye system. In the 2014 World Volleyball Championship, the system has also been introduced. In the U.S. Open in August last year, except Arthur Ashe and Armstrong still using the traditional manual lineup, the other 15 courts were all introduced with hawk-eye live broadcast system for the first time. There was only one referee in the competition field, which had minimized the number of staff in the court.

While the hawk-eye system is increasingly popular for sports competitions, the existing ELC systems are based on binocular vision technologies, which still have many problems. For example, the installation and maintenance process of the existing ELC systems are very complex (need to conduct court measurement, 3D modeling, and camera calibration); The costs of installation, operation and management are very high with multiple high-speed cameras and synchronization of cameras. These problems prevent the wider use of the ELC systems for sport games.

Recently in the research work of TTNet, Myint, Kotera, etc. [2-9] it was reported that monocular vision could be used to detect and track fast moving objects, which could be applied to the ELC system. We are motivated to solve the problems of ELC associated with binocular vision. With the significant advances on computer vision, we believe it is possible to develop a simple, effective and cheap ELC system with monocular vision. In this paper, we investigate the feasibility of monocular vision based ELC system, and successfully propose a monocular vision technology based ELC method. We develop and implement the ELC method, test it with custom datasets. Experiments proves the feasibility of the method, which can be used to produce effective and cheap ELC systems.

### 1.2  Related Works

**Research on Hawk-Eye system based on Binocular Vision:** Most of the existing ELC systems are based on binocular vision technology. Paper [1,10] describes the basic principles and key technologies of the hawk-eye system. Paper [11] is an US patent about the hawk-eye system. It also describes the process of the ELC. Paper [12] implemented a hawk-eye system based on binocular vision. Paper [13] implemented a hawk-eye system based on LabVIEW and MATLAB.

According to the above research, the hawk-eye system, based on the principles of binocular vision and camera calibration technology, predicts the bouncing point to determine whether the ball is out of bounds according to the trajectory of the ball. More details are shown in Figure. 1.

**Binocular vision.** Binocular vision is the method based on the parallax (the direction difference caused by observing the same target from two different points with a certain distance) in order to obtain the three-dimensional geometric information of an object from multiple images.

**Camera calibration.** In machine vision, the transformation relationship between the spatial position of an object in the world three-dimensional coordinate system and the two-dimensional position in the image coordinate system needs to be determined by the camera calibration parameters (internal and external parameters). The process of camera calibration is to solve the internal and external parameters through the relationship between 3D and 2D coordinates.

The binocular vision based hawk-eye systems have the following major processing steps: 1) Set up multiple high-speed cameras in different locations, calibrate the camera to obtain the internal and external parameters of each camera. 2) Measure the court and perform a virtual reconstruction of the court, store the relevant data of the court and rules; 3) Identify the spherical target in each frame of the picture, get the location of the ball. 4) Use the ball position information from at least two different cameras at the same time to calculate the three-dimensional world coordinates of the ball. 5) Obtain the three-dimensional motion trajectory of the ball according to the relationship of consecutive frames, predict the actual landing position. 6) Map the trajectory of the ball and the predicted bouncing point to the virtually reconstructed court to identify whether the ball is out of bounds or not.

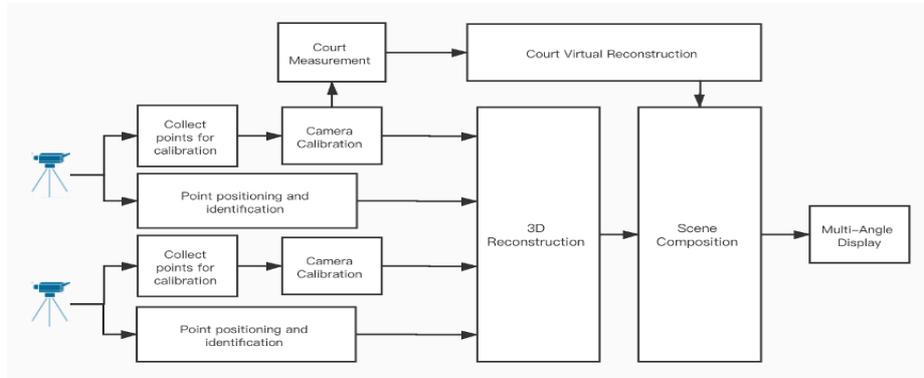

**Figure. 1.** The building blocks of hawk-eye system.

### 1.3    Our Contributions

Compared to the existing works, our work made the following major contributions.

1) We designed and implemented a ELC method for tennis based on monocular vision. According to our investigation, there are no published papers, patents or other documents applying monocular vision to tennis hawk-eye system, which shows innovation of our work

2) We designed a two-stage tennis positioning method that combines background subtraction and color area filtering, which can solve the problem of misjudgment and missing judgments in the monocular vision based ELC system.

3) We designed a bouncing point prediction method based on the minimum fitting loss of uncertain points, which can solve the problem of unobvious changing trend near the bouncing point.

4) We collected and marked 349 video samples, implemented the core algorithm and prototype system, including tennis trajectory positioning, bouncing point prediction, out-of-bounds judgment, etc. We carried out the experimental verification and analyzed the experimental results, which demonstrated the feasibility of the proposed low cost and effective ELC method.

## 2    Methodology

### 2.1    Overall processes.

In this section, we present the methodology of the proposed ELC system design. The overall process of the proposed ELC method includes the following major steps:

(1) Identify each tennis ball in each frame of the video;

(2) Fit the two curves: falling trend and rising trend;

(3) Calculate the intersection point of two curves as the prediction of the bouncing point;

(4) Determine whether the ball is out of bounds or not by comparing the relative position of the bouncing point and the court lines.

Figure. 2. shows the results of our program after each step being visualized. The detailed processes of the method are shown in Figure. 3. Compared with the traditional binocular vision hawk-eye system (Figure. 1), our method does not need camera calibration and 3D coordinate reconstruction. Instead, we directly use the relative relationship of 2D coordinate to determine whether it is out of bounds or not. The following part of this section will focus on the tennis ball positioning method and the bouncing point prediction method.

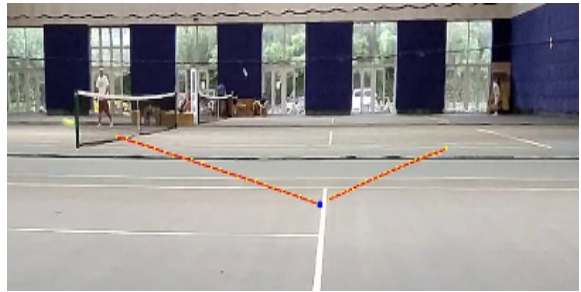

**Figure. 2.** The yellow parts are the point of tennis balls identified, the red parts are two fitting curves (falling and rising), the blue parts are the intersection point of two curves, which is the predicted bounce point. We deterimine whether the ball is out of bound or not by the relative position of the bouncing point and the court lines.

### 2.2    Tennis Positioning Method

Common methods [14-18] used in detecting high-speed moving objects in video include optical flow [16], frame difference [17], and background subtraction [14, 15].

**Optical flow method.** The change of pixel intensity in time domain and the neighborhood correlation of pixels is used to determine the pixel motion, and the change of pixels between adjacent frames is compared to determine the target position and find the motion information. Optical flow method can detect moving objects without knowing any information of the scene, but noise, multiple light sources, shadows and occlusion will seriously affect the calculation results of optical flow field distribution. Moreover, the calculation of optical flow method is complex, and it is difficult to realize real-time processing.

**Frame difference method.** frame difference method is a non parametric modeling method, which distinguishes moving foreground and background by pixel comparison. It selects the images which are continuous in time or separated by a fixed number of frames for difference. By selecting an appropriate threshold, the pixels in the difference image are divided into moving foreground and background. However, the frame difference method is sensitive to environmental noise, and the selection of threshold is very important. If the threshold is too low, it is not enough to suppress the noise, and if it is too high, it is easy to ignore the changes in the image. For large and uniform color moving objects, there may be holes in the interior of the object, which can not extract the moving object completely. It can only be used when the camera is still, but there are many limitations in its application.

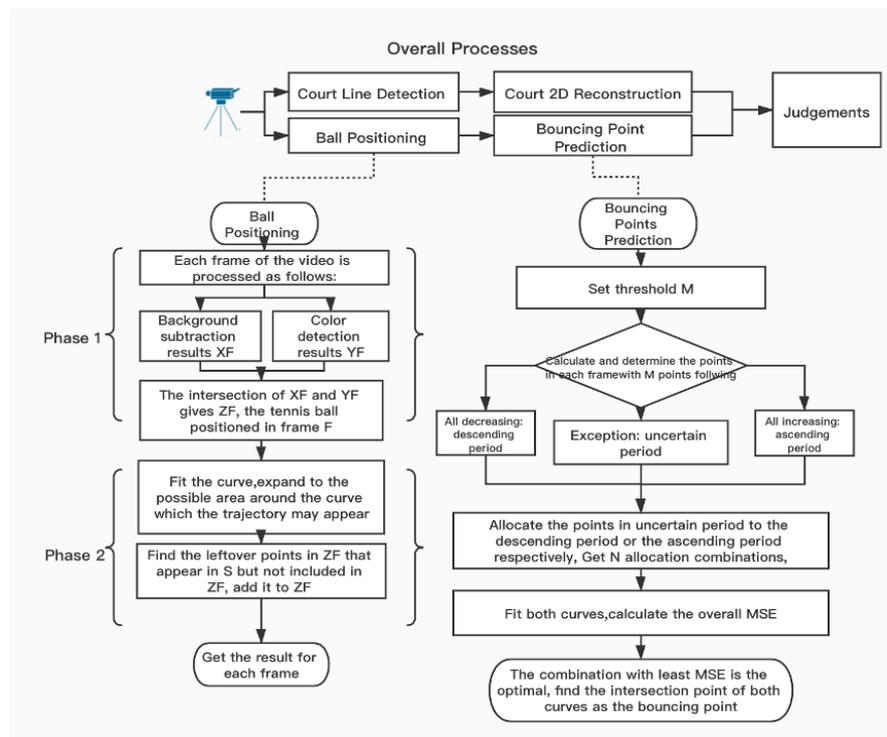

**Figure. 3.** Detailed process on our ELC method.

**Background subtraction method.** The basic idea of background subtraction is to get the approximate background image by using the background parameter model, and then compare the current frame with the approximate background image by difference. The area with larger difference is considered as the moving area, and the area with smaller difference is the background area. In this method, background modeling and updating are crucial, because the background image must be updated in real time according to the illumination and external environment changes. Aiming at the problem of scene dynamic change, researchers have proposed many modeling algorithms, such as MOG Gaussian mixture model separation algorithm [14] and KNN background separation method [15].

### 2.3    Tennis Bouncing Point Prediction Method

**Related Works.** In the hawk-eye system based on binocular vision, the prediction of bouncing point is mainly realized by the method of 3D coordinate reconstruction. The is not reported work on predicting bouncing point based on monocular vision. Bouncing point detection is based on the observed bounce of the ball. When the bouncing occurs, the ball's trajectory suddenly changes. We design an algorithm to detect the mutation to realize the bouncing point detection. TTNet treated the bounce of the ball as an event to learn, events also include service, the ball hits the net, etc. By building specific sequential events such as deep learning network, TTNet could detect different events. In paper [8], Rozumnyi et al.(2019) propopsed a method to detect trajectory mutation by using dynamic programming to find global energy minimization and the bouncing points.

**Our Method.** We found that there are several challenges exist in the bouncing point prediction methods: as mentioned above, the bounce of the ball is a mutation. It is relatively easy for us to distinguish the descend before bouncing and the ascend after bouncing. However, for a part of the balls appear just before or after the bouncing point, because of the speed, rotation, the change of the recording angle, the bounce of the ball trajectory mutation is not obvious. This makes it hard to distinguish between whether the ball is descending or ascending.

We propose a bouncing point prediction method based on the minimum fitting loss of an uncertain point. The process of this method is shown in Figure. 3. The general idea of this method is described as follows. Our goal is to divide the trajectory of the ball into two categories: descending phase and ascending phase. By using the least square method, both curves are fitted. The intersection point of the two curves can be regarded as the predicted bouncing point. As some of the balls are too difficult to distinguish between whether it is descending or ascending, we first mark these balls as uncertainty (as shown in Figure. 4-a), three orange points on the bottom are uncertain points), then put these uncertain points in descending phase or ascending phase respectively and get a variety of combinations. Fit both curves with the least square method, find the combination which MSE (minimum fitting error) is the smallest, that distribution of descending and ascending points is the final result. The intersection point of two curves are the final bouncing point. We use an open-source least-square curve fitting code [19] on GitHub for our work. Figure. 4-b is the result of our algorithm fitting.

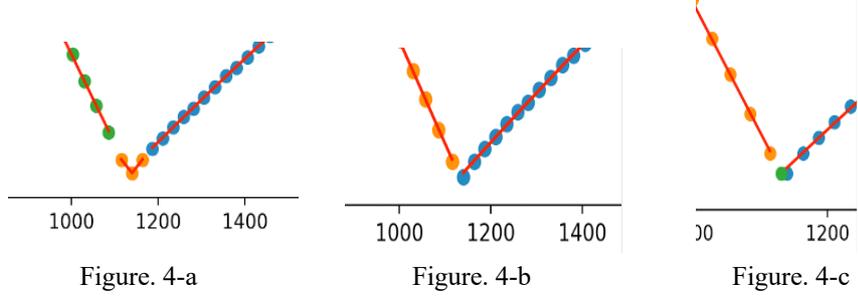

Figure. 4-a            Figure. 4-b            Figure. 4-c

**Figure. 4.** Representative results of our fitting algorithm. a) The uncertainty point is near the falling place (three orange points on the bottom) ; b) The result of the optimal fitting error; c) The result of the bouncing point (intersection point) in green.

## 3    Experiment

### 3.1    Datasets

In this section, we present the development of dataset for our experiments and the experiment results. In table 1, we show our sample quantity and other added information. The datasets are already public on GitHub, but we have removed the link due to blind review.

Table 1. Sample quantity and other added information

| Total number of data | 349 |
|---|---|
| The Number of Confusing data | 11 |
| Proportion of confusing data | 3.2% |
| Average time(s) | 10.0 |

We chose Samsung S10 mobile slow-motion (240 frames per second) to capture videos in this project. The shooting device is fixed, which can cover the service line and the baseline on one side. Figure. 5 is a representative sample of the confusing data.

### 3.2    Results & Discussion

**Results.** In our experiemnts, we evaluate the accuracy of the judgment of balls out of bound or not in the samples. The formula of accuracy is the number of video samples judged correctly divided by total number of video samples. The judging method is calculated manually based on the relationship of location between the predicted bouncing point and the marked lines of the court.

$$L_v(\mathrm{M}(X_v), t) = 1_{\{D(M(X_v),t)<\varepsilon\}} \quad (1)$$

Monocular Visual Analysis For Electronic Line Calling of Tennis Games    9

$$R_{suc} = \sum_i \frac{||L_v(M(X_{v,i}), t_i)||}{N} \qquad (2)$$

We developed and implemented the proposed ELC system. After experiments, we compute the accuracy of the method over the video samples (see table 2).

**Table 1.** Accuracy of the ELC system

|  | Number | Success | $R_{suc}$ |
|---|---|---|---|
| Normal | 338 | 336 | 99.4% |
| Confusing | 11 | 9 | 81.8% |
| Total | 349 | 345 | 98.9% |

One of the errors is shown in Figure. 5. The main reason is that there is strong interference and the court line is far away from the camera.

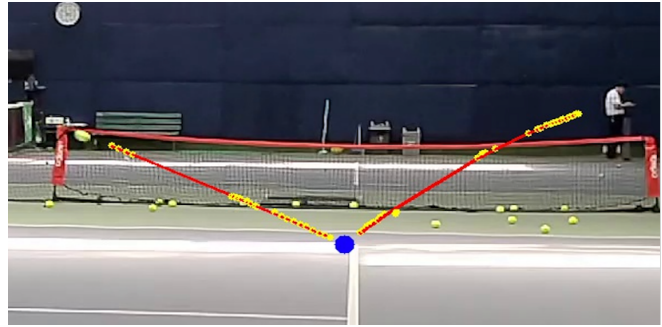

**Figure. 5.** A representative misjudgement on confusing segments

**Influence on the accuracy of the camera position.** Next we evaluate the influence of camera position on the accuracy. Three images are presented in Figure. 6. It can be seen that it is relatively easy to judge image Figure. 6-a, but the other images are difficult to judge even if we watch the video for multiple times (see Figure 6-b and 6-c ). This may be explained by that the monocular vision method relies on the relative position rather than 3D coordinate reconstruction. Different camera positions will cause parallax, thus resulting in lower accuracy. Generally speaking, we think that four cameras should be placed on court (two cameras at each side of the court). One in between the middle of the service line and baseline, and the other is in the middle of the sideline. Finding the optimal arrangement of these cameras is left for our future work.

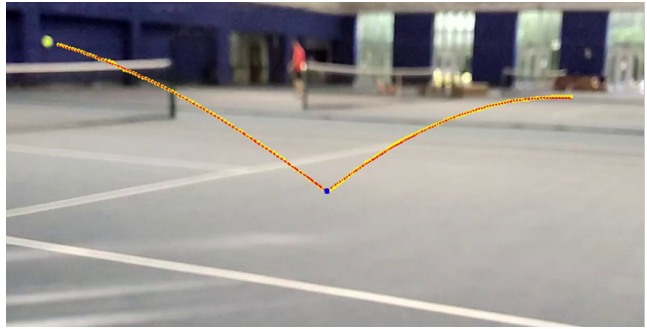

Figure. 6-a

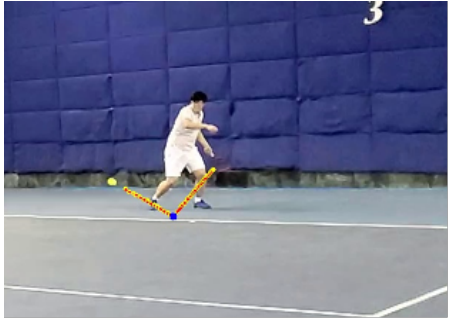    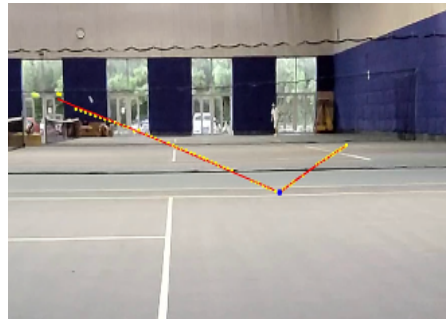

Figure. 6-b                           Figure. 6-c

**Figure. 6.** Examples of balls with different placement. a) In the middle of the court; b) Near the baseline of the court; c) Near the far sideline of the court.

## 4      Conclusion

In order to solve the problems in the binocular vision based hawk-eye system such as high cost and complicated installation and maintenance, we designed a monocular vision based method. With this method we first identify the tennis balls in the video and predict the bounce point.Then according to the 2D image coordinates and tennis court lines we can judge whether the ball is out of line or not. In order to solve the problem of missed judgment and misjudgment in positioning, we designed a two-stage tennis positioning method that combines background subtraction and color area filtering methods. Aiming at the problem that the changing trend of the bounce point is unobvious, we designed a bouncing point prediction method based on the minimum fitting loss of the uncertainty point.

In order to verify our method, we collected and annotated 349 video samples, and implemented the core algorithm, including tennis ball trajectory positioning, bouncing point prediction, etc. We conducted expermients and the results showed that the proposed method is feasible and effective. It has an accuracy of 99.4% for the normal scenarios and an accuracy of 81.8% for challenging scenarios.

To the best of our knowledge, there is no work or attempt reported on monocular vision based ELC system for tennis games, which prove the innovation of our work. We have shared on Github [20] all the annotated samples and codes of this research work.